\documentclass[runningheads]{llncs}
\usepackage[T1]{fontenc}
\usepackage{graphicx,verbatim}
\usepackage{amsmath,amssymb}
\usepackage{booktabs}
\usepackage{multirow}
\usepackage{comment}
\usepackage[noend]{algpseudocode}
\usepackage{algorithm}
\usepackage{hyperref}
\hypersetup{
    colorlinks=true,
    linkcolor=blue,
    citecolor=blue,
    urlcolor=blue,
    }
    
\usepackage[table]{xcolor}
\definecolor{LightCyan}{rgb}{0.88,1,1}
\usepackage{eso-pic}

\AddToShipoutPictureFG{%
  \AtPageLowerLeft{%
    \raisebox{0.5cm}{%
      \makebox[\paperwidth]{%
        \footnotesize\textit{Accepted at MICCAI 2026}%
      }%
    }%
  }%
}

\begin{document}

\title{Unsupervised Brain Anomaly Detection as a Bayesian Inverse Problem with Diffusion Prior}

\author{Hugues Roy\inst{1}\orcidID{0009-0006-3503-6849} \and
Reuben Dorent\inst{1}\textsuperscript{*}\orcidID{0000-0002-7530-0644} \and
Ninon Burgos\inst{1}\textsuperscript{*}\orcidID{0000-0002-4668-2006}}


\authorrunning{H. Roy et al.}

\institute{
Sorbonne Université, Institut du Cerveau - Paris Brain Institute - ICM, CNRS, Inria, Inserm, AP-HP, Hôpital de la Pitié Salpêtrière, F-75013, Paris, France\\
\textsuperscript{*}Contributed equally \\
\email{huguesroy1277@gmail.com}}

\titlerunning{UAD as Bayesian Inverse Problem with Diffusion Prior}

\maketitle              

\begin{abstract}

\sloppy Unsupervised anomaly detection (UAD) aims to localize abnormal regions in medical scans without pixel-level annotations. A typical strategy seeks to reconstruct a pseudo-healthy image that preserves subject-specific anatomy. Recently, diffusion models have been proposed to perform UAD. However, these methods rely on heuristic noise schedules or synthetic corruptions to balance subject-specificity and anomaly removal.
In this work, we propose an alternative formulation of UAD as a Bayesian inverse problem under a diffusion prior. First, we introduce a latent spatial anomaly mask that models pixel-wise consistency between a test image and its latent corresponding pseudo-healthy image. Then, we propose an approximation of the unknown generation process that links healthy anatomy, anomalies, and the observed image, enabling a well-defined likelihood within the Bayesian framework. Building on recent advances in diffusion-based inverse problem methods, we jointly infer the pseudo-healthy image and the anomaly mask via annealed posterior sampling. We evaluate our approach on FDG PET (ADNI) and FLAIR MRI (BraTS 2021), demonstrating improved anomaly localization performance compared to other diffusion-based approaches and validating the contribution of our introduced model. Our code is available at \url{https://github.com/HuguesRoy/UAD_DAPS}.

\keywords{Unsupervised anomaly detection  
\and Inverse problem}

\end{abstract}

\section{Introduction}

Unsupervised anomaly detection (UAD) in medical imaging aims to identify abnormal regions without requiring assumptions about anomaly characteristics or pixel-level annotations. A common strategy is to reconstruct a pseudo-healthy version of an abnormal image to detect anomalies.

Such reconstruction must satisfy two competing objectives: (i) it should be plausible under the healthy data distribution, and (ii) it should preserve subject-specific anatomical details.

Diffusion models trained on healthy data have recently been used for UAD due to their strong generative capabilities. However, to balance these competing objectives, these methods rely on heuristic choices, such as denoising noise levels~\cite{behrendt_Patched_2024,bercea_Mask_2023a,bercea_Diffusion_2024,wolleb_Diffusion_2022}, corruption strategies~\cite{bercea_Diffusion_2024,wolleb_Binary_2024,wyatt_AnoDDPM_2022,Roy2025AnoBFN}, or synthetic anomalies~\cite{kascenas_Denoising_2022,liang_iter_2024,bei_madad_2026}.

In parallel, diffusion models recently achieved strong performance in Bayesian inverse problems by leveraging learned data priors \cite{song_PseudoinverseGuided_2022,zhang_Improving_2025a}. These approaches progressively estimate a latent clean signal from noisy observations through iterative denoising updates. However, they rely on a known forward model linking the clean signal to the noisy measurement, which is unavailable in UAD.

In this work, we bridge these two lines of research by formulating unsupervised anomaly detection as a Bayesian inverse problem in which the forward model is unknown and must be approximated.
Our contributions are fourfold: (1) we formulate diffusion-based UAD as a Bayesian inverse problem with a latent anomaly mask; (2) we introduce an approximation of the unknown forward model; (3) we propose a spatially regularized mask prior enabling stable joint inference via intermediate posterior sampling \cite{zhang_Improving_2025a}; and (4) we conduct extensive experiments demonstrating the effectiveness of our approach which outperforms existing diffusion-based methods on both PET and MRI benchmarks.

\section{Inverse problem using diffusion models}

Inverse problems aim to reconstruct an unknown signal $\boldsymbol{x}_0$ from a partial, noisy measurement $\boldsymbol{y}$ by inverting a known \emph{forward model} that describes the measurement process. In a Bayesian setting, reconstruction is formulated as inference under the posterior distribution $p(\boldsymbol{x}_0 \mid \boldsymbol{y}) \propto p(\boldsymbol{y} \mid \boldsymbol{x}_0) p(\boldsymbol{x}_0)$, where the likelihood $p(\boldsymbol{y} \mid \boldsymbol{x}_0)$ is defined by the forward model and $p(\boldsymbol{x}_0)$ is the prior distribution. A key challenge of inverse problems is to approximate this unknown prior. 

Recently, diffusion models have shown strong performance to approximate priors~\cite{cardoso_Monte_2023,chung_Diffusion_2022,dou_DiffSamp_2024,kawar2022denoising,song_PseudoinverseGuided_2022} in inverse problems. A noisy signal $\boldsymbol{x}_t$ is iteratively denoised using both a prior diffusion model trained on clean signals and the likelihood $p(\boldsymbol{y} \mid \boldsymbol{x}_t)$, leading to a denoising process conditioned on the measurement $\boldsymbol{y}$.
However, early denoising errors have been found to be difficult to correct~\cite{zhang_Improving_2025a}. 

To address this limitation, Zhang et al.~\cite{zhang_Improving_2025a} proposed the Decoupled Annealed Posterior Sampling (DAPS)  strategy. Rather than strictly following the diffusion trajectory, DAPS introduces a sequence of intermediate posteriors $p_k(\boldsymbol{x}_0\mid \boldsymbol{y}, \boldsymbol{x}_{t_k}) \propto p(\boldsymbol{y}\mid\boldsymbol{x}_0) p_k(\boldsymbol{x}_0 \mid \boldsymbol{x}_{t_k})$ which gradually converge to the true posterior $p(\boldsymbol{x}_0 \mid \boldsymbol{y})$ and enable exploring a larger solution space. As in \cite{chung_Diffusion_2022,song_PseudoinverseGuided_2022}, they approximate the conditional prior $p_k(\boldsymbol{x}_0 \mid \boldsymbol{x}_{t_k})$ using a Gaussian distribution, 
\begin{equation}
p_k(\boldsymbol{x}_0 \mid \boldsymbol{x}_{t_k})
\approx \mathcal{N}\left(
\boldsymbol{x}_0;
\hat{\boldsymbol{x}}_0(\boldsymbol{x}_{t_k}, t_k), r_k^2 \boldsymbol{I}
\right),
\label{eq:diffusion_prior_gaussian}
\end{equation}
where $\hat{\boldsymbol{x}}_0$ is sampled from the reverse process of the diffusion model. The variance $r_k^2$ weights the contribution of the conditional prior in the intermediate posterior.

Instead of directly sampling $\boldsymbol{x}_{t_{k+1}}$ from $\boldsymbol{x}_{t_k}$, an intermediate signal $\boldsymbol{x}_0^{(k)}$ is sampled using the intermediate posterior $p_k(\boldsymbol{x}_0\mid \boldsymbol{y}, \boldsymbol{x}_{t_k}) $ and re-noised to obtain a sample $\boldsymbol{x}_{t_{k+1}}$. Thanks to this decoupled sampling strategy and intermediate posterior parameterization, larger changes between iterations are obtained, preventing early denoising errors from propagating through the sampling process.

A key limitation of DAPS and other inverse problem approaches is their reliance on a known and available forward model. This assumption typically does not hold in unsupervised anomaly detection for medical imaging, where no forward model exists to sample an abnormal image $\boldsymbol{y}$ from a healthy image $\boldsymbol{x}_0$.

\section{Method}
 
In this work, we propose a new diffusion-based approach to unsupervised anomaly detection. Specifically, we formulate anomaly detection as a Bayesian inverse problem and approximate the unknown forward model by introducing a latent variable $\boldsymbol{m}$ that represents a probabilistic anomaly mask. Under this formulation, the objective is to jointly infer the healthy image $\boldsymbol{x}_0$ and the mask $\boldsymbol{m}$ from the observed image $\boldsymbol{y}$. In the following sections, we define the probabilistic model, including the likelihood, the priors, and the proposed inference strategy.

\begin{figure}[t]
    \centering
    \includegraphics[width=\linewidth]{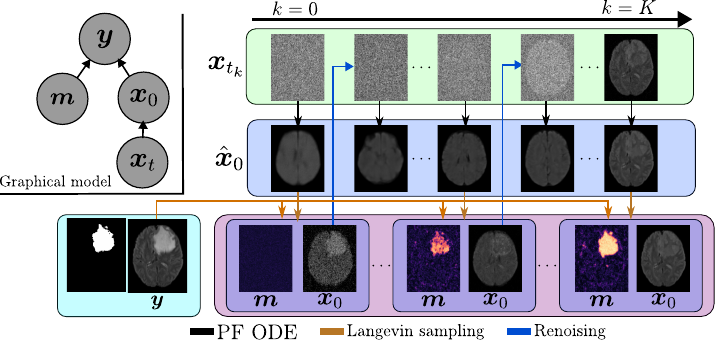}
    \caption{\textbf{Top left:} Probabilistic graphical model of the proposed method.
    \textbf{Right:} Overview of the inference process. As $k$ increases, the posterior distribution progressively concentrates, enabling the sampling of the final estimates $\boldsymbol{x}_0$ and $\boldsymbol{m}$. 
    The variable $\boldsymbol{x}_{t_k}$ denotes the noisy version of $\boldsymbol{x}_0$ at step $k$, which allows a decoupling between the previously sampled noisy variable and the newly generated one.}
    \label{fig:method}
\end{figure}

\subsubsection{Generative model}

To formalize anomaly detection as a Bayesian inverse problem,
we assume that the observed image $\boldsymbol{y}$ is generated from a latent healthy image $\boldsymbol{x}_0$ and a latent anomaly mask $\boldsymbol{m}$ through the likelihood $p(\boldsymbol{y} \mid \boldsymbol{x}_0, \boldsymbol{m})$. Specifically, 
the observation $\boldsymbol{y}$ is expected to coincide with $\boldsymbol{x}_0$
 only in regions labeled as normal by the mask $\boldsymbol{m}$. As shown in the graphical model in Fig.~\ref{fig:method}, assuming prior independence between $\boldsymbol{x}_0$ and $\boldsymbol{m}$, the joint distribution factorizes as $p(\boldsymbol{x}_0, \boldsymbol{m}, \boldsymbol{y}) = p(\boldsymbol{x}_0) p(\boldsymbol{m}) p(\boldsymbol{y} \mid \boldsymbol{x}_0, \boldsymbol{m}),$ where $p(\boldsymbol{x}_0)$ is the prior over healthy images and $p(\boldsymbol{m})$ is a spatial prior over anomaly masks. As in Bayesian inverse problems, the objective is to sample the healthy image $\boldsymbol{x}_0$ and anomaly mask $\boldsymbol{m}$ from the posterior $p(\boldsymbol{x}_0, \boldsymbol{m} \mid \boldsymbol{y}) \propto p(\boldsymbol{y} \mid \boldsymbol{x}_0, \boldsymbol{m}) p(\boldsymbol{x}_0) p(\boldsymbol{m})$. To approximate this posterior, we adopt a similar formulation as in  DAPS~\cite{zhang_Improving_2025a}. Introducing the noisy diffusion states $\boldsymbol{x}_{t_k}$ to approximate the image prior $p(\boldsymbol{x}_0)$, we define a sequence of intermediate posteriors
\begin{align}
     p_k(\boldsymbol{x}_0, \boldsymbol{m} \mid \boldsymbol{y}, \boldsymbol{x}_{t_k})
     \propto
     p_k(\boldsymbol{y} \mid \boldsymbol{x}_0 , \boldsymbol{m})\,
     p_k(\boldsymbol{x}_0 \mid \boldsymbol{x}_{t_k})\,
     p(\boldsymbol{m}),
     \label{eq:daps_uad_posterior}
\end{align}
where $p_k(\boldsymbol{x}_0 \mid \boldsymbol{x}_{t_k})$ is the conditional diffusion prior defined in Eq.~\eqref{eq:diffusion_prior_gaussian}.

\subsubsection{Approximate the forward model}

Unlike classical inverse problems, the forward model relating a healthy image $\boldsymbol{x}_0$ to an abnormal scan $\boldsymbol{y}$ is unknown. In this section, we propose an approximation of the forward model $p(\boldsymbol{y} \mid \boldsymbol{x}_0, \boldsymbol{m})$ based on the introduced anomaly mask $\boldsymbol{m}$.
The probabilistic anomaly mask $\boldsymbol{m}$ encodes pixel-level abnormality: values close to 0 indicate anomalous measurements, whereas values close to 1 enforce consistency between $\boldsymbol{y}$ and $\boldsymbol{x}_0$. Accordingly, we model the likelihood as:
\begin{align}
    p_k(\boldsymbol{y} \mid \boldsymbol{x}_0, \boldsymbol{m}) =
    \mathcal{N} (\boldsymbol{y};
        \boldsymbol{x}_0,
        r_k^2 \, \mathrm{diag}(\boldsymbol{m})^{-2}) \enspace ,
        \label{eq:approximate_likelihood}
\end{align}
with variance $r_k^2$, matched to the conditional prior 
$p_k(\boldsymbol{x}_0 \mid \boldsymbol{x}_{t_k})$ variance. 
This ensures that, at intermediate step $k$, in normal regions, the variability of the diffusion-based prediction is consistent with the variability observed in $\boldsymbol{y}$. Deviations exceeding this expected variability cannot be explained by the prior and are therefore attributed to abnormal regions.

\subsubsection{Spatial prior on the mask}

Since the probabilistic mask $\boldsymbol{m}$ is introduced as a latent variable, we must specify a prior distribution $p(\boldsymbol{m})$.
To enforce values in $(0,1)$, we parametrize it as $\boldsymbol{m} = s(\boldsymbol{a})$, where $\boldsymbol{a}$ are unconstrained logits and $s$ denotes the sigmoid function.
We then define a prior  $p(\boldsymbol{a})$, that (i) defines the initial mask value, (ii) controls the magnitude of mask updates during inference, and (iii) enforces spatial consistency. To this end, we assume a Gaussian Markov random field prior:
\begin{align}
    p(\boldsymbol{a}) = \mathcal N(\boldsymbol{a}; \mu \boldsymbol{1}, \boldsymbol{Q}^{-1}) \enspace ,
    \label{eq:prior_mask}
\end{align}
with a precision matrix $\boldsymbol{Q} = \lambda_0 \boldsymbol{I} + \lambda_c \boldsymbol{L}$, where $\lambda_0, \lambda_c >0$ and $\boldsymbol{L}$ is the graph Laplacian of the image grid. The quadratic form $\boldsymbol{a}^\top \boldsymbol{L} \boldsymbol{a}$ penalizes differences between neighboring logits, promoting spatial consistency. The parameter $\mu$ sets the initial value of the mask, $\lambda_0$ controls the global shrinkage of logits toward $\mu$, and $\lambda_c$ governs the spatial consistency strength.

\subsubsection{Energy formulation of the intermediate posterior}

Finally, by combining the conditional diffusion prior defined in Eq.~\eqref{eq:diffusion_prior_gaussian}, the approximate likelihood in Eq.~\eqref{eq:approximate_likelihood} and the logit mask prior in Eq.~\eqref{eq:prior_mask}, the intermediate posteriors in Eq.~\eqref{eq:daps_uad_posterior} can be formulated as a Gibbs distribution, i.e. $ p_k(\boldsymbol{x}_0,\boldsymbol{a} \mid \boldsymbol{y}, \boldsymbol{x}_{t_k}) \propto \exp (-U_k(\boldsymbol{x}_0,\boldsymbol{a}))$, where $U_k(\boldsymbol{x}_0,\boldsymbol{a})$ is defined as:
\begin{align}
    \begin{aligned}
        U_k(\boldsymbol{x}_0,\boldsymbol{a})
        &=
        \underbrace{\frac{1}{2r_k^2}
        \big\|\boldsymbol{x}_0-\hat{\boldsymbol{x}}_0(\boldsymbol{x}_{t_k},t_k)\big\|^2}_{\text{Conditional prior consistency }}
        +
        \underbrace{\frac{1}{2r_k^2}
        \big\|s(\boldsymbol{a})\odot(\boldsymbol{y}-\boldsymbol{x}_0)\big\|^2}_{\text{Mask-weighted measurement consistency}}
         \\
        &\qquad  \underbrace{- \sum_i \log s(a_i)
        + \frac{\lambda_0}{2}\|\boldsymbol{a}-\mu \boldsymbol{1} \|^2}_{\text{Mask value regularization}}
        + \underbrace{\frac{\lambda_c}{2}
        \sum_{(i,j)\in E}(a_i-a_j)^2}_{\text{Spatial mask consistency}} .
    \end{aligned}
    \label{eq:energy}
\end{align}

\subsubsection{Inference by Langevin dynamics}

In order to sample from the intermediate posteriors, we use unadjusted Langevin dynamics \cite{well_langevin_2011} targeting the energy $U_k$ at each intermediate step $k$. We denote $j$ the Langevin steps and $(\boldsymbol{x}_0^{(j)}, \boldsymbol{a}^{(j)})$ the samples at step $j$. At each step $k$, the reconstruction $\boldsymbol{x}_0$ is initialized using the mean of the conditional prior sample and the observation. The mask logits are initialized once at $k=0$ as $\boldsymbol{a}^{0} = \mu \boldsymbol{1}$ and are propagated across intermediate steps without re-initialization, enabling progressive refinement of the anomaly mask as the noise level decreases.
\begin{align}
    \boldsymbol{x}_0^{(j+1)}
    &=
    \boldsymbol{x}_0^{(j)}
    - \eta_x \nabla_{\boldsymbol{x}_0} U_k(\boldsymbol{x}_0^{(j)}, \boldsymbol{a}^{(j)})
    + \sqrt{2\eta_x}\,\boldsymbol{\xi}_x^{(j)}, \\
    \boldsymbol{a}^{(j+1)}
    &=
    \boldsymbol{a}^{(j)}
    - \eta_a \nabla_{\boldsymbol{a}} U_k(\boldsymbol{x}_0^{(j)}, \boldsymbol{a}^{(j)})
    + \sqrt{2\eta_a}\,\boldsymbol{\xi}_a^{(j)},
\end{align}
where $\boldsymbol{\xi}_x^{(j)}$ and $\boldsymbol{\xi}_a^{(j)}$ are standard Gaussian noise terms. We use only a few iterations per intermediate step $k$. The step sizes $\eta_x, \eta_a >0$ are automatically adjusted based on gradient magnitudes $\| \nabla_{\boldsymbol{x}_0} U_k(\boldsymbol{x}_0^{(j)}, \boldsymbol{a}^{(j)})\|^2_2$ and $\| \nabla_{\boldsymbol{a}} U_k(\boldsymbol{x}_0^{(j)}, \boldsymbol{a}^{(j)})\|^2_2$.

\subsubsection{Hyperparameter reparameterization}

The posterior energy in Eq.~\eqref{eq:energy} depends explicitly on the variance schedule $r_k$. We parametrize $r_k = c\,\sigma_{t_k}$, where $\sigma_{t_k}$ denotes the diffusion noise level and $c>0$ is a calibration constant estimated on a healthy validation set to ensure stable reconstruction in the absence of anomalies. As $r_k$ decreases, the relative strength of the likelihood increases compared to the mask prior since it scales as $1/r_k^2$. If the prior weight $\lambda_0$ is kept fixed across intermediate steps, this changing scale alters the balance between the likelihood and the logit prior.
To preserve a stable trade-off across intermediate steps, we adapt $\lambda_0$. It is chosen such that the gradient magnitudes from the likelihood and mask value are almost equal.

\begin{figure}[!tbh]
    \centering
    \includegraphics[width=0.9\linewidth]{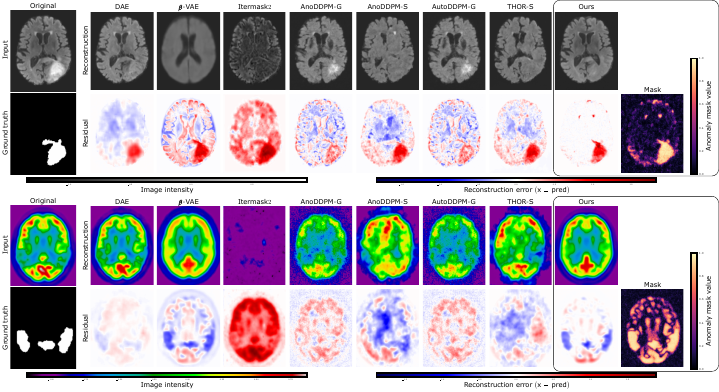}
    \caption{Qualitative results on BraTS FLAIR (top) and ADNI PET (AD 50\%, bottom).}
    \label{fig:results}
\end{figure}

\subsubsection{Implementation details}

We adopt the diffusion parameterization of Karras et al.~\cite{karras_Elucidating_2022a} and train the diffusion model exclusively on healthy data. The conditional estimate $\hat{\boldsymbol{x}}_0(\boldsymbol{x}_{t_k}, t_k)$ in  Eq.~\eqref{eq:diffusion_prior_gaussian} is obtained using the probability flow ordinary differential equation \cite{song_ScoreBased_2020} (PF-ODE) with two solver steps per noise level. For all experiments, we use 150 intermediate posterior steps in the DAPS framework and 75 Langevin iterations at each intermediate step. The noise levels $\{t_k\}$ follow the DAPS parameterization with $\sigma_{\max} = 20$. We set the spatial consistency parameters $\lambda_c$ to 1. An 80GB NVIDIA H100 GPU was used for training.

\section{Experiments \& Results}

To evaluate the performance of our proposed model, experiments were conducted on 2D FLAIR MRI of glioblastoma patients and on 2D FDG PET with synthetic Alzheimer-like hypometabolism. Our framework was compared with UAD diffusion models \cite{bercea_Mask_2023a,bercea_Diffusion_2024,wyatt_AnoDDPM_2022} and state of the art methods \cite{Hassanaly2025Benchmarking,kascenas_Denoising_2022,liang_iter_2024}. For the proposed approach, we use the inferred latent anomaly mask directly as anomaly maps, while other methods use residuals.

\noindent\textbf{Datasets}~
\textbf{ADNI PET}
FDG PET images were obtained from ADNI \cite{petersenAlzheimersDiseaseNeuroimaging2010} and preprocessed with Clinica~\cite{routier_clinica_2021} (affine registration to MNI, intensity normalization, cropping, resampling to $128^3$, rescaling to $[-1,1]$). 
We selected 621 images from 301 cognitively normal subjects. 
Fifty images were kept for testing. 
Because no ground truth exists, we use the framework described in \cite{Hassanaly2024EvaluationPseudohealthy} to generate synthetic images with Alzheimer-like hypometabolism (50\%). The remaining images were split at the subject level into training (545) and validation (25 volumes) sets. 
\textbf{BRATS FLAIR}
FLAIR images from BraTS 2021 \cite{Menze_brats_2015} and ADNI \cite{petersenAlzheimersDiseaseNeuroimaging2010} were preprocessed with Clinica \cite{routier_clinica_2021} in a similar manner as the PET images. Models were trained on healthy ADNI FLAIR slices (234 subjects) and evaluated on BraTS slices containing tumors (1,251 subjects). 
To analyze sensitivity to anomaly size, slices were grouped into medium and large categories using the 33\textsuperscript{rd} and 66\textsuperscript{th} percentiles of lesion area. One slice per subject was randomly selected per category, yielding 819 large lesions and 1,156 medium lesions.

\noindent\textbf{Evaluation}
For the PET images, because multiple slices come from the same subject, we account for this hierarchical structure using stratified bootstrapping at the subject level when computing metrics. We also report the inference time. Detection performance was evaluated using the best possible Dice score ($\lceil \text{Dice} \rceil$) and pixel-wise average precision (AP). Significance was assessed via Wilcoxon signed-rank tests with Bonferroni correction for multiple comparisons ($p<0.01$).

\noindent\textbf{Baselines} We compare our method to diffusion-based UAD approaches, including \textit{AnoDDPM} \cite{wyatt_AnoDDPM_2022} (G-Gaussian and S-simplex noise variants), \textit{AutoDDPM} (Gaussian noise) \cite{bercea_Mask_2023a}, and \textit{THOR} (Simplex noise) \cite{bercea_Diffusion_2024}. For fairness, all diffusion models use the same U-Net backbone. We additionally compare against \textit{$\text{IterMask}^2$} \cite{liang_iter_2024}, a recent state-of-the-art UAD framework that uses synthetic anomalies, and a variational autoencoder ($\beta$-\textit{VAE}) \cite{Hassanaly2025Benchmarking,Loizillon2024Detecting}. All models are trained on healthy data only. Hyperparameters were tuned on the healthy validation set.

\begin{table}[tb]
\caption{Performance on BraTS FLAIR. Best possible Dice and AP are reported in \% as mean (standard deviation). Arrows indicate favorable direction of each metric.}
\label{table:brats}
\centering
\small
\setlength{\tabcolsep}{4pt}
\renewcommand{\arraystretch}{1}
\begin{tabular}{llcccc}
\toprule
& 
& \multicolumn{2}{c}{\textbf{Large}} 
& \multicolumn{2}{c}{\textbf{Medium}} \\
\cmidrule(lr){3-4} \cmidrule(lr){5-6}
\textbf{Category} & \textbf{Method} 
& $\lceil \text{Dice} \rceil$(\%) $\uparrow$ & AP (\%) $\uparrow$ 
& $\lceil \text{Dice} \rceil$ (\%) $\uparrow$ & AP (\%) $\uparrow$ \\
\midrule

\multirow{3}{*}{Reconstruction}
& DAE \cite{kascenas_Denoising_2022}
& $68.6 (17.5)$ & $71.9 (20.5)$
& $66.9 (21.0)$ & $69.0 (25.5)$ \\

& $\beta$-VAE \cite{Loizillon2024Detecting}
& $56.3 (16.5)$ & $54.2 (22.4)$
& $48.3 (19.8)$ & $44.7 (25.4)$ \\

& IterMask2 \cite{liang_iter_2024}
& $\mathbf{78.7 (14.5)}$ & $\mathbf{82.3 (17.0)}$
& $\mathbf{71.0 (18.3)}$ & $\mathbf{72.8 (22.3)}$ \\

\midrule

\multirow{5}{*}{Diffusion}
& AnoDDPM-G \cite{wyatt_AnoDDPM_2022}
& $36.7 (9.8)$ & $30.3 (11.5)$
& $31.8 (13.9)$ & $24.5 (15.3)$ \\

& AnoDDPM-S \cite{wyatt_AnoDDPM_2022}
& $63.1 (19.6)$ & $64.9 (24.5)$ 
& $51.5 (23.5)$ & $50.1 (29.0)$ \\

& AutoDDPM-G \cite{bercea_Mask_2023a}
& $44.7 (19.4)$ & $39.1 (21.5)$
& $45.2 (22.7)$ & $40.6 (26.6)$ \\

& THOR-S \cite{bercea_Diffusion_2024}
& $62.6 (20.4)$ & $60.4 (26.9)$ 
& $50.3 (21.6)$ & $42.5 (26.5)$ \\

\cmidrule(lr){2-6}
& \textbf{Ours}
& $72.3 (19.2)$ & $73.0 (24.3)$
& $63.8 (22.6)$ & $62.5 (28.3)$ \\

\bottomrule
\end{tabular}
\end{table}

\begin{table}[tb]
\caption{Performance on ADNI PET. Mean $\lceil \text{Dice} \rceil$ and AP are reported in \% with 95\% bootstrap confidence intervals. Inference speed is reported as slices per second.}
\label{tab:adni}
\centering
\small
\setlength{\tabcolsep}{4pt}
\renewcommand{\arraystretch}{1}

\begin{tabular}{llccc}
\toprule
& & \multicolumn{3}{c}{\textbf{Synthetic Alzheimer-like hypometabolism}} \\
\cmidrule(lr){3-5}
\textbf{Category} & \textbf{Method} & $\lceil \text{Dice} \rceil$ (\%) $\uparrow$ & AP (\%) $\uparrow$ & Slices/s $\uparrow$ \\
\midrule

\multirow{3}{*}{Reconstruction}
& DAE \cite{kascenas_Denoising_2022}
& $47.3 [46.0, 48.6]$ & $46.8 [44.7, 48.8]$ & $296.7$ \\

& $\beta$-VAE \cite{Hassanaly2025Benchmarking}
& $61.9 [61.2, 62.6]$ & $64.5 [63.4, 65.7]$ & $\mathbf{535.4}$ \\

& IterMask2 \cite{liang_iter_2024}
& $40.1 [39.7, 40.6]$ & $17.0 [16.7, 17.3]$ & $108.7$ \\

\midrule

\multirow{5}{*}{Diffusion}
& AnoDDPM-G \cite{wyatt_AnoDDPM_2022}
& $25.2 [24.9, 25.4]$ & $14.1 [14.0, 14.3]$ & $2.1$ \\

& AnoDDPM-S \cite{wyatt_AnoDDPM_2022}
& $40.9 [39.3, 42.5]$ & $34.7 [32.4, 37.3]$ & $2.5$ \\

& AutoDDPM-G \cite{bercea_Mask_2023a}
& $9.8 [9.3, 10.3]$ & $13.7 [13.6, 13.9]$ & $1.2$ \\

& THOR-S \cite{bercea_Diffusion_2024}
& $43.0 [41.6, 44.4]$ & $35.0 [33.3, 36.7]$ & $2.1$ \\

\cmidrule(lr){2-5}

& \textbf{Ours}
& $\mathbf{65.7 [64.7, 66.7]}$ 
& $\mathbf{67.9 [66.2, 69.7]}$ 
& $2.8$ \\

\bottomrule
\end{tabular}
\end{table}

\noindent\textbf{Results} Quantitative results on the BraTS dataset are reported in Table~\ref{table:brats}. Diffusion-based approaches (\textit{AnoDDPM}, \textit{AutoDDPM}, \textit{THOR}) achieve lower AP and $\lceil \text{Dice} \rceil$ compared to reconstruction-based methods relying on synthetic anomalies (\textit{$\text{IterMask}^2$}) or single-step denoising (\textit{DAE}). Among diffusion models, \textit{Ours} shows the best performance ($p<0.01$) for both lesion sizes and both evaluation metrics. Compared to all methods, it ranks second on large lesions and third on medium lesions. From Fig.~\ref{fig:results}, we observe that the \textit{DAE} and \textit{$\text{IterMask}^2$} attenuate abnormal hyperintensities toward normal tissue values. In contrast, \textit{Ours} shows restoration of the anomaly, for example reconstructing the ventricles. On ADNI PET (Table~\ref{tab:adni}), our method outperforms all baselines by $+3.8$pp in $\lceil \text{Dice} \rceil$ and $+3.4$pp in AP ($p<0.01$). We observe that \textit{$\text{IterMask}^2$}, which performs strongly on BraTS, fails to generalize to this modality. Finally, inference speed shows that our method does not introduce computational overload compared to standard diffusion models while achieving better detection performance.

\noindent\textbf{Ablation study} We evaluate the contribution of each modeling component on the BraTS dataset (Table~\ref{tab:ablation}). \textbf{(A) No latent mask.} Removing the latent anomaly mask results in no detection; the abnormal image is reconstructed. \textbf{(B) Fixed likelihood variance ($r_k^2$).} 
Instead of using the variance schedule, the likelihood variance is fixed to $5\times10^{-2}$ as commonly done in classical inverse problems~\cite{zhang_Improving_2025a}. This modification degrades performance.
\textbf{(C) No spatial regularization.} Removing spatial consistency in the mask prior (i.e., setting $\lambda_c = 0$) negatively impacts anomaly detection performance. This highlights the importance of spatial regularization to enforce coherent anomaly regions and suppress isolated false positives.

\begin{table}[t]
\centering
\caption{Ablation experiments on BraTS FLAIR (large lesions only). Best possible Dice and AP are reported in \% as mean (standard deviation).}
\label{tab:ablation}
\small
\setlength{\tabcolsep}{4pt}
\renewcommand{\arraystretch}{1}

\begin{tabular}{lcc}
\toprule
\textbf{Method} 
& $\lceil \text{Dice} \rceil$ (\%) $\uparrow$ 
& AP (\%) $\uparrow$ \\
\midrule

(A) No latent mask
&$0.0 (0.0)$ & $7.9 (2.2)$ \\

(B) Fixed likelihood variance
&$65.8 (19.3)$ & $64.6 (25.6)$ \\

(C) No spatial regularization
&$46.6 (13.9)$ & $38.1 (15.4)$ \\

\midrule
\textbf{Ours}
& $\mathbf{72.3 (19.2)}$ & $\mathbf{73.0 (24.3)}$ \\

\bottomrule
\end{tabular}
\end{table}

\section{Conclusion} 

In summary, we formulated unsupervised anomaly detection as a Bayesian inverse problem, using a diffusion model as a prior over healthy anatomy. Our method introduces a latent anomaly mask that is inferred jointly with the reconstruction through sampling of a sequence of intermediate posteriors. The mask provides an approximation of the unknown forward model, while a spatial prior ensures coherent and anatomically consistent anomaly regions. Ablation studies confirmed the importance of each proposed component. Experimentally, our model achieves competitive performance on BraTS and state-of-the-art results on ADNI PET compared to multiple baselines.

While the current formulation already demonstrates competitive performance, it also opens several directions for further development. Replacing the Gaussian approximation of the conditional diffusion prior with a more expressive model could improve the approximation of the intermediate posteriors and enhance anomaly detection. Similarly, refining the forward model formulation may allow the framework to better capture pathological patterns. Since our method relies on joint intermediate posterior inference, alternative sampling or optimization strategies may improve stability and computational efficiency. Extending the framework to full 3D volumes is another important direction. Together, these improvements would extend the framework to other imaging modalities and clinical problems.

\begin{credits}

\subsubsection{\ackname}
This work was supported by ANR through the ``Investissements d’avenir'' program (ANR-10-IAIHU-06, ANR-19-P3IA-0001, PRAIRIE 3IA Institute), the ``France 2030'' program (ANR-23-IACL-0008, PRAIRIE-PSAI), and the MediTwin project. It also received support from the European Union’s Horizon Europe Framework Programme under grant agreement No. 101136607 (CLARA), and from Next Generation EU through the ``France Relance'' plan. The ARAMIS Lab is affiliated with DIM C-BRAINS, funded by the Conseil Régional d’Ile-de-France. Experiments used GENCI--IDRIS HPC resources under grants 2025-AD011011648R5 and AD011017165. R.D. received Marie Skłodowska-Curie grant No. 101154248 (SafeREG). Data collection and sharing were funded by the Alzheimer’s Disease Neuroimaging Initiative (ADNI), supported by NIH grant U19AG024904, with additional public and private-sector support.

\subsubsection{\discintname}The authors have no competing interests to declare that are relevant to the content of this article.

\end{credits}

\bibliographystyle{splncs04}
\bibliography{paper-4982}

\end{document}